\documentclass[letterpaper, 10 pt, journal, twoside]{IEEEtran}
\usepackage{amsmath,amsfonts}
\usepackage[linesnumbered,ruled,vlined]{algorithm2e}
\usepackage{array}
\usepackage{textcomp}
\usepackage{stfloats}
\usepackage{url}
\usepackage{verbatim}
\usepackage{graphicx}
\def\BibTeX{{\rm B\kern-.05em{\sc i\kern-.025em b}\kern-.08em
		T\kern-.1667em\lower.7ex\hbox{E}\kern-.125emX}}

\usepackage{balance}

\usepackage{color}
\definecolor{darkred}{RGB}{205,38,38}

\usepackage{hyperref}
\hypersetup{
	colorlinks=true,
	linkcolor=darkred, 
	filecolor=darkred,      
	urlcolor=red,
	citecolor=blue, 
}

\usepackage{amsthm,amssymb}
\usepackage{mathrsfs}
\usepackage{caption}
\usepackage{subfig}
\usepackage{float} 
\usepackage{enumitem}
\usepackage{booktabs} 
\usepackage{multirow}
\usepackage{url}
\usepackage{braket}
\usepackage{cite}
\usepackage{flushend}

\usepackage{array}

\newcolumntype{L}[1]{>{\raggedright\let\newline\\\arraybackslash\hspace{0pt}}m{#1}}
\newcolumntype{C}[1]{>{\centering\let\newline\\\arraybackslash\hspace{0pt}}m{#1}}
\newcolumntype{R}[1]{>{\raggedleft\let\newline\\\arraybackslash\hspace{0pt}}m{#1}}

\usepackage{color,soul}
\usepackage{threeparttable}
\usepackage{makecell}

\soulregister\cite7
\soulregister\ref7 

\soulregister\citep7 
\soulregister\citet7 
\soulregister\pageref7 

\usepackage[dvipsnames, svgnames, x11names]{xcolor} 
\usepackage{xpatch}
\makeatletter
\ExplSyntaxOn
\cs_new:Npn \bibColoredItems #1#2
{
	\clist_map_inline:nn {#2} { \cs_new:cpn {bib@colored@##1} {#1} } 
}
\ExplSyntaxOff

\newcommand\bib@setcolor[1]{%
	\ifcsname bib@colored@#1\endcsname
	\expanded{\noexpand\color{\csname bib@colored@#1\endcsname}}%
	\else
	\normalcolor
	\fi
}

\IfPackageLoadedTF{hyperref}{\@tempswatrue}{\@tempswafalse}
\if@tempswa
\xpatchcmd\@bibitem {\H@item}{\bib@setcolor{#1}\H@item}{}{\PatchFailed}
\xpatchcmd\@lbibitem{\H@item}{\bib@setcolor{#2}\H@item}{}{\PatchFailed}
\else
\xpatchcmd\@bibitem {\item}  {\bib@setcolor{#1}\item}  {}{\PatchFailed}
\xpatchcmd\@lbibitem{\item}  {\bib@setcolor{#2}\item}  {}{\PatchFailed}
\fi
\makeatother

\begin{document}

\title{
Multi-Robot Rendezvous in Unknown Environment with Limited Communication
}

\author{Kun Song, Gaoming Chen, Wenhang Liu, and Zhenhua Xiong, \IEEEmembership{Member, IEEE}
\thanks{Manuscript received: May, 13, 2024; Revised August, 1, 2024; Accepted August, 27, 2024.}
\thanks{
	This paper was recommended for publication by Editor M. Ani Hsieh upon evaluation of the Associate Editor and Reviewers' comments.
}
\thanks{
	This work was in part supported by the National Natural Science Foundation of China (U1813224) and MoE Key Lab of Artificial Intelligence, AI Institute, Shanghai Jiao Tong University, China. \textit{(Corresponding author: Zhenhua Xiong.)}
}
\thanks{
	The authors are with the School of Mechanical Engineering, State Key Laboratory of Mechanical System and Vibration, Shanghai Jiao Tong University, Shanghai, China (e-mail: coldtea@sjtu.edu.cn; cgm1015@sjtu.edu.cn; liuwenhang@sjtu.edu.cn; mexiong@sjtu.edu.cn).
}
\thanks{Digital Object Identifier (DOI): see top of this page.}
}

\markboth{IEEE Robotics and Automation Letters. Preprint Version. Accepted August, 2024}
{SONG \MakeLowercase{\textit{et al.}}: Multi-Robot Rendezvous in Unknown Environment with Limited Communication}
\maketitle

\begin{abstract}
Rendezvous aims at gathering all robots at a specific location, which is an important collaborative behavior for multi-robot systems.
However, in an unknown environment, it is challenging to achieve rendezvous.
Previous researches mainly focus on special scenarios where communication is not allowed and each robot executes a random searching strategy, which is highly time-consuming, especially in large-scale environments.
In this work, we focus on rendezvous in unknown environments where communication is available. 
We divide this task into two steps: rendezvous based environment exploration with relative pose (RP) estimation and rendezvous point selection.
A new strategy called partitioned and incomplete exploration for rendezvous (PIER) is proposed to efficiently explore the unknown environment, where lightweight topological maps are constructed and shared among robots for RP estimation with very few communications.
Then, a rendezvous point selection algorithm based on the merged topological map is proposed for efficient rendezvous for multi-robot systems.
The effectiveness of the proposed methods is validated in both simulations and real-world experiments.
\end{abstract}


\section{Introduction}
Multi-robot systems have been receiving increasing attention in recent years \cite{zhou2023racer,park2017fault}.
The cooperative motion in multi-robot systems has been widely studied, for example, consensus \cite{olfati2004consensus}, exploration \cite{zhou2023racer}, and rendezvous \cite{ando1999distributed}.
Among the above-mentioned cooperative motion, rendezvous is one of the most important topics, which aims at gathering all robots at a specific location at a given moment \cite{alpern1995rendezvous}.
This behavior is common in the real world, for example, a few friends want to meet with each other in a shopping mall or a rescue helicopter is looking for lost hikers in the desert \cite{ozsoyeller2021multi}.

The rendezvous problem has been studied for a long time\cite{ando1999distributed, ozsoyeller2013symmetric,park2017fault, li2022rendezvous,jang2021multirobot,mathew2015multirobot}.
Currently, previous research can be classified into two types based on whether communication is allowed.
For the first scenario, communication between robots is not allowed \cite{ozsoyeller2013symmetric,ozsoyeller2022m,ozsoyeller2019rendezvous,weber2012optimal}.
In these researches, robots are assumed to have limited observation range and the rendezvous is achieved when the robot is in the observation range of other robots.
Due to that the environment is unknown initially and communication is not allowed, each robot always performs a random search strategy to achieve rendezvous.

One of the simplest rendezvous problems involves two robots moving along a straight line \cite{alpern1995rendezvous}.
The solutions to this problem can be divided into two types: symmetric and asymmetric.
The symmetric version requires the robots to execute the same strategy.
Symmetric rendezvous strategy for two robots switching among three fixed locations was proved to be optimal in \cite{weber2012optimal}.
Besides, it was extended to a continuous scenario in \cite{ozsoyeller2013symmetric}.
The asymmetric version was studied in \cite{leone2022search,meghjani2014asymmetric}.
Rendezvous for $n > 2$ robots on a line was studied in \cite{ozsoyeller2021multi} and was extended to 2D scenario in \cite{ozsoyeller2019rendezvous,ozsoyeller2022m}.
In these researches, robots rely on stochastic strategies for rendezvous.
However, in scenarios with a large number of robots or large-scale environments, the random searching strategy makes rendezvous challenging or even infeasible.

The condition of communication was introduced in \cite{lopez2012visual,ando1999distributed,ganguli2009multirobot,park2017fault,park2018robust}, where the environment is usually assumed to be known and robots can locate themselves in the environment.
Moreover, RPs between robots are available under this assumption.
This kind of rendezvous problem has been receiving significant attention in control theory.
However, rendezvous in unknown environments is more common and valuable.
In real world, the bandwidth is always limited. 
For instance, satellite-based communication in the wilderness or weak signals in underground parking, their bandwidth can even be less than 1\,Mbps.
This limitations can lead to time delays in the system, thereby affecting control and decision-making.
Therefore, it is important to study rendezvous under the assumption of limited bandwidth communication.
As far as we know, this is the first work to perform rendezvous under this assumption.

Recently, multi-robot exploration with unknown initial \textbf{r}elative \textbf{p}oses (RP) has been studied \cite{tian2022kimera, zhou2023racer, zhang2022mr}.
In the process, robots determine their RPs and merge the constructed map.
In theory, after the exploration and map construction, it is possible to use this full map to achieve rendezvous.
However, simply applying multi-robot exploration algorithms to achieve rendezvous is inefficient.
Firstly, the exploration aims to construct a complete and detailed map.
However, for rendezvous, exploring the environment as minimally as possible is preferred to improve the efficiency, which means an incomplete exploration is needed.
Secondly, what kinds of information should be transmitted has also not been studied to effectively achieve RP estimation and rendezvous point selection, which will lead to high communication cost.

Thus, we propose a strategy called partitioned and incomplete exploration for rendezvous (PIER).
We focus on how to effectively utilize the limited communication bandwidth (we assume communication range is not limited).
This problem involves two aspects specifically: first, how to achieve RP estimation, and second, how to transmit environmental structural information to enable the selection of rendezvous points.
In our previous paper, we proposed a lightweight \textbf{F}eature-based \textbf{H}ybrid \textbf{T}opological \textbf{Map} (FHT-Map) \cite{song2023fht}.
There are two types of nodes: main nodes and support nodes.
Main nodes store features extracted from the environment and can facilitate RP estimation.
Environmental structural information is stored in support nodes with very low storage requirements, which can be used for rendezvous point selection.
We implement the lightweight FHT-Map in this new rendezvous task.
By sharing these Maps among robots, we can achieve rendezvous with low bandwidth requirement.

Thus, we divide the rendezvous problem into two steps.
Firstly, each robot performs rendezvous based exploration using PIER and constructs its FHT-Map. 
During this process, robots share the maps among their neighbors and detect similar main nodes for RP estimation, and the FHT-Maps can be merged one by one.
Secondly, once the RPs between all robots are determined, the optimal rendezvous point can be selected on the merged map.
Our main contributions are:

\begin{itemize}[leftmargin=*]
	\item A novel framework is proposed for solving rendezvous problem in unknown environment with limited communication.
	\item PIER is proposed to perform efficient incomplete exploration with unknown initial RPs.
	\item Lightweight FHT-Map is used for rendezvous, enabling efficient RP estimation and rendezvous point selection.
\end{itemize}

\section{Preliminaries}
In this paper, we consider the rendezvous problem in a finite 2D space $S \subset \mathbb{R}^2$, where the size of the environment is unknown initially.
This assumption is easy to satisfy \cite{roy2001collaborative}, for example, in rendezvous problems for humans within a shopping mall or a campus.
We define the k-th robot in the multi-robot system as $\mathcal{R}^k$ and the total number of robots is $n$.
$\mathbf{P}^k=[\mathbf{p}^k,\mathbf{q}^k] \in \mathbf{SE}(2)$ describes the pose of $\mathcal{R}^k$ in its own map frame, where $\mathbf{p}^k \in \mathbb{R}^2$ and $\mathbf{q}^k \in \mathbf{SO}(2)$ denote the position and orientation of $\mathcal{R}^k$.
Besides, a distributed communication framework is used, where all communication are point-to-point and occurs between two robots without a central server.
An undirected, unweighted, connected, and non-time-varying communication graph $\mathcal{G}_c=(\mathcal{V}_c,\mathcal{E}_c)$ is used to describe the communication topology between $n$ robots.
The node set is $\mathcal{V}_c=\{\mathcal{R}_1,\cdots,\mathcal{R}_n\}$, the edge set is $\mathcal{E}_c \subseteq \mathcal{V}_c \times \mathcal{V}_c$.
$\mathcal{N}_c^k$ defines the set of neighboring robots on $\mathcal{G}_c$ for $\mathcal{R}^k$.

In this work, each robot maintains a lightweight FHT-Map.
Both features and structural information of the environment are stored in the map.
The FHT-Map of $\mathcal{R}^k$ can be represented by an undirected graph $\mathcal{G}_{f}^k = (\mathcal{V}_f^k,\mathcal{E}_f^k)$, where $\mathcal{V}_f^k$ represents the set of nodes and $\mathcal{E}_f^k$ represents the set of edges. 
Nodes are categorized into two types: $\mathcal{V}_{f,\text{main}}^k \subset \mathcal{V}_f^k$ refers to main nodes, and $\mathcal{V}_{f,\text{sup}}^k\ \subset \mathcal{V}_f^k$ is known as support nodes.
Each node $v_f^{k,i}$ is indexed with an id $i$.
The contents of main nodes and support nodes are different.
Main nodes contain all information presented in Table \ref{table:node_info}.
Support nodes only contain position $\textbf{p}^{k,i}$ and rectangular free space $s^{k,i}$.
Each edge $e_f = (v_f^{k,m},v_f^{k,n})\in \mathcal{E}_f^k$ connects two nodes with id $m$ and $n$, and represents a traversable path between the two nodes in free space.

\begin{table}[t]
	\renewcommand{\arraystretch}{1.4}
	\caption{Contents Stored in a Node}
	\label{table:node_info}
	\centering
	\begin{tabular}{|l||l|}
		\hline
		$\textbf{p}^{k,i} \in \mathbb{R}^{2}$ & position of node $i$ in map frame of $\mathcal{R}^k$\\
		\hline
		$s^{k,i} \in \mathbb{R}^{4}$ & rectangular local free space around node $i$\\
		\hline
		$\boldsymbol{\phi}^{k,i} \in \mathbb{R}^{512}$ & feature vector extracted from RGB images at $p^{k,i}$\\
		\hline
		$\boldsymbol{\psi}^{k,i} \in \mathbb{R}^{360} $ & local laser scan at $p^{k,i}$\\
		\hline
	\end{tabular}
\end{table}

Different data can be used for RP estimation, for example, images and point clouds.
Using raw data from the sensors for RP estimation provides more accurate results but imposes a significant burden on communication. 
In this paper, we employ lightweight Place Recognition (PR) descriptors \cite{milford2012seqslam,yin2023automerge,garg2022semantic} for rough RP estimation.
Only when two PR descriptors are matched, 2D laser scans of these two points will be transferred for a more accurate RP estimation.
A time-varying, undirected and unweighted RP graph $\mathcal{G}_r=(\mathcal{V}_r,\mathcal{E}_r)$ is used to describe the existence of estimated RPs among $n$ robots.
The node set in $\mathcal{G}_r$ is $\mathcal{V}_r=\{\mathcal{R}^1,\cdots,\mathcal{R}^n\}$, the edge set is $\mathcal{E}_r \subseteq \mathcal{V}_r \times \mathcal{V}_r$.
Initially, since there are no estimated RPs between any two robots, $\mathcal{E}_r$ is an empty set.
When a RP is obtained between robots $m$ and $n$, the edge $(\mathcal{R}^m,\mathcal{R}^n)$ is added to $\mathcal{E}_r$.
Besides, the relative poses between any two robots located on the same subgraph can be obtained through  Pose Graph Optimization (PGO), which will be further described in Section \ref{sec:RPEst}.

When $\mathcal{G}_r$ becomes a connected graph, meaning the RPs between all robots are determined, the selection of the rendezvous point is performed.
In \cite{litus2008distributed,zebrowski2007energy}, the rendezvous point is the point where the sum of costs for all robots to move to that point is minimized.
However, this assumption is based on minimizing the energy required for rendezvous, while in this paper, we consider achieving it as quickly as possible.
The details for the selection of the rendezvous point will be further described in Section \ref{sec:OptRend}.

\begin{figure*}[!t]
	\centering
	\includegraphics[width=7in]{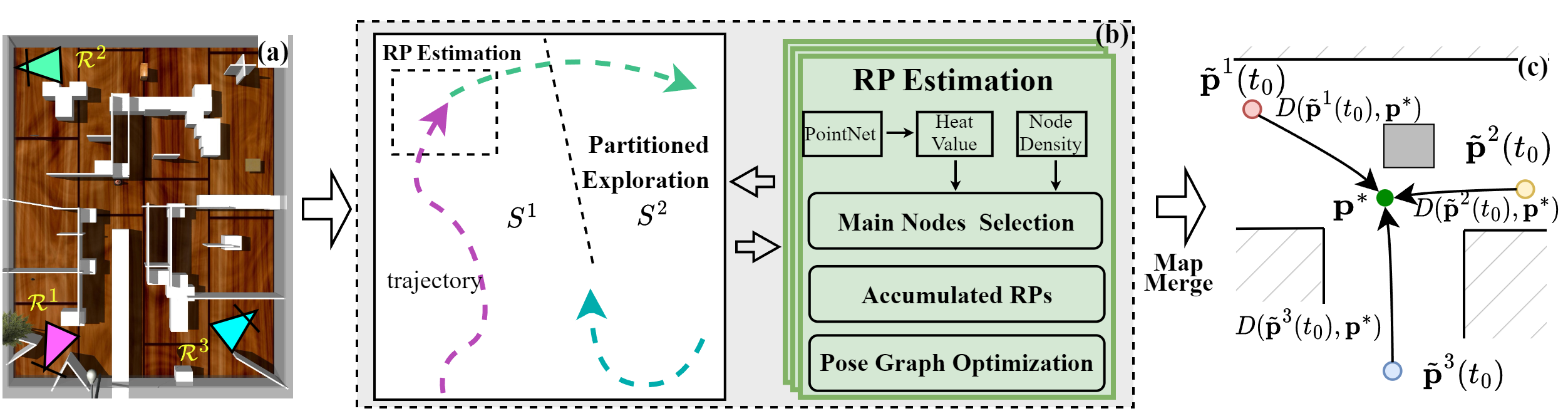}
	\caption{The framework for robots rendezvous used in this work. (a) Three robots are located in different locations. (b) PIER is performed firstly to explore the environment. During this process, each robot maintains a FHT-Map and calculates RPs. (c) When $\mathcal{G}_r$ is connected, rendezvous point is selected based on the merged FHT-Map and rendezvous is achieved.}
	\label{fig:total}
\end{figure*}

The framework for robots rendezvous is shown in Fig. \ref{fig:total}, which divides the task into two steps: PIER exploration with RP estimation, and rendezvous point selection. 

\section{PIER Formulation}
\label{sec:PIER}
Robot exploration for a complete reconstruction of the environment has been widely investigated in recent years \cite{zhang2022mr,hardouin2023multirobot,zhou2023racer,bi2023cure}.
In \cite{zhang2022mr}, Next Best View (NBV) is used and each robot employs a greedy strategy, enabling rapid acquisition of environmental features but potentially resulting in poor cooperation between robots. 
In \cite{hardouin2023multirobot,zhou2023racer}, multi-robot exploration is modeled as a multiple traveling salesman problem (m-TSP), which optimizes the movements based on current frontiers.
However, m-TSP-based methods are not efficient for the unknown space exploration.
Compared with a detailed reconstruction, rendezvous based exploration aims at acquiring environmental features for RP estimation.
Therefore, utilizing a greedy strategy to acquire features and considering cooperation between robots will be time-efficient for rendezvous.
So, \textbf{p}artitioned and \textbf{i}ncomplete \textbf{e}xploration for \textbf{r}endezvous (PIER) is proposed.
In PIER, we combine NBV-based and partitioned strategies to achieve rapid cooperative exploration by multiple robots for rendezvous under the condition of unknown initial RPs.

\subsection{Single Robot NBV Exploration}
The NBV-based method employs a greedy strategy, which can achieve a faster coverage for incomplete exploration tasks.
During this process, the environment $S$ can be divided into two parts: known space $S_\text{known}$ and unknown space $S_\text{unknown}$.
Furthermore, it can also be classified into two types: free space $S_\text{free}$ and occupied space $S_\text{occ}$.
In PIER, $\mathcal{R}^k$ is assigned with a portion of the area $S^k$ to explore.
The combination of each space forms the environment $S^1 \cup S^2 \cdots \cup S^n = S$.

Frontiers are defined as the boundaries between $S_\text{free}$ and $S_\text{unknown}$.
Frontiers in $S^k$ are described using $\{f^k_i|i=1,2,3,\dots\}$.
Through clustering, we obtain the clustered centers of $\{f^k_i\}$ denoted as $\{\hat{f}^k_i\}$, while filtering out the cluster centers with little number of frontiers.
We define the navigation cost from $\textbf{P}^k$ to $\hat{f}^k_i$ as $\textbf{C} (\textbf{P}^k, \hat{f}^k_i)$, which is obtained on topological maps.
The frontier center with minimal cost will be chosen as the next goal for $\mathcal{R}^k$, which is the NBV-based strategy for a single robot.

\subsection{RP Estimation between Robots}
\label{sec:RPEst}
To achieve rendezvous, each robot first performs NBV-based strategy to explore the unknown environment. 
During this process, each robot constructs an FHT-Map \cite{song2023fht} which can be used for RP estimation.
In the FHT-Map, there are two types of nodes: main nodes and support nodes. 
Main nodes store the visual features $\boldsymbol{\phi}^{k,i} \in \mathbb{R}^{512}$ and geometric features $\boldsymbol{\psi}^{k,i}\in \mathbb{R}^{360}$ of the location, which can be used for RP estimation. 
When $\mathcal{R}^k$ passes a main node of its neighboring robots $\mathcal{N}_c^k$, it can obtain the RP between $\mathcal{R}^k$ and that robot.
So, it is crucial to select appropriate positions of main nodes to accelerate RP estimation and rendezvous.
In this section, we will describe the details of using FHT-Maps for RP estimation.

\subsubsection{Selection of Main Nodes}
Given a fixed number of main nodes, these nodes should be distributed as uniformly as possible to increase the possibility of being visited by others.
Assuming the current FHT-Map of $\mathcal{R}^k$ is $\mathcal{G}^k_f$, we define the main nodes' density at $\textbf{p}^k(t)$ as
\begin{equation}
	\mathcal{D}(\textbf{p}^k(t)) = \sum_{v_f^{k,i}\in \mathcal{V}_{f,\text{main}}^k} \exp(-\frac{||\textbf{p}^k(t) - \textbf{p}^{k,i}||_2^2}{\sigma_c^2})
\end{equation}
where $\sigma_c$ is a hyper-parameter that controls the density of main nodes.
$\textbf{p}^{k,i}$ is the position of a constructed main node in the FHT-Map of $\mathcal{R}^k$.
When $\mathcal{R}^k$ moves towards unknown space, $\mathcal{D}(\textbf{p}^k(t))$ will decrease.
Therefore, given a series of candidate main nodes in the robot's trajectory, we tend to choose the one with smaller $\mathcal{D}(\textbf{p}^k(t))$ to create the main node.

\begin{figure}[!t]
	\centering
	\includegraphics[width=3in]{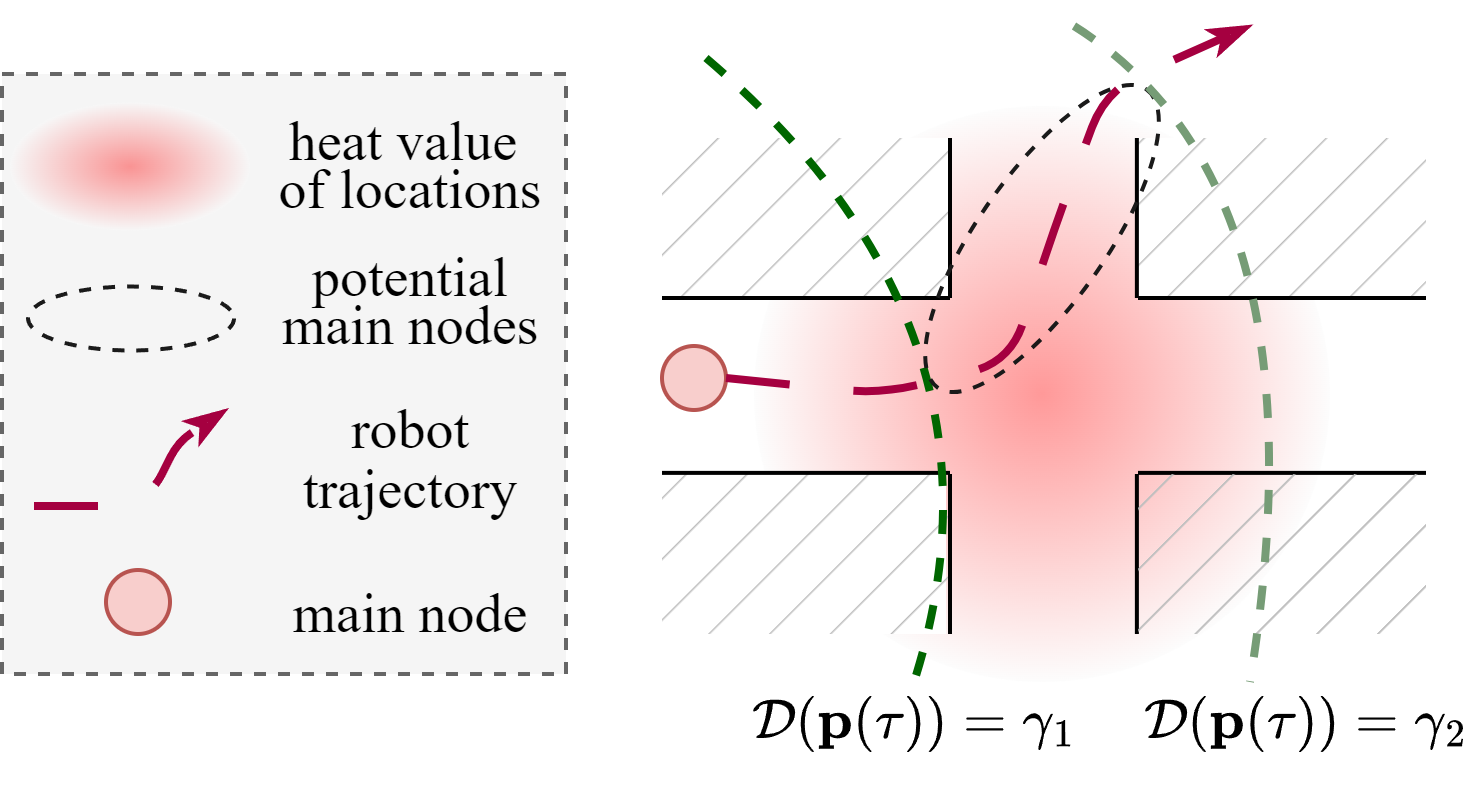}
	\caption{
		Illustration for main nodes selection. The robot's motion trajectory passes a crossroad which has higher probability of being visited by other robots.
	}
	\label{fig:main_node_select}
\end{figure}

Secondly, since each robot performs incomplete exploration, the probability of each position being visited is different. 
An intuitive scenario is that robots are more likely to pass through crossroads, which is shown in Fig. \ref{fig:main_node_select}. 
Therefore, selecting main nodes at these positions can facilitate RP estimation.
We refer to the probability of a location $\textbf{p}^k(t)$ being visited as ``heat value'' $\mathcal{H}(\textbf{p}^k(t)) \in \mathbb{R}$.
We use a neural network to fit the heat value for each location. 
This network takes the laser scan $\boldsymbol{\psi}(\textbf{p}^k(t))$ as input and outputs the fitted $\mathcal{H}(\textbf{p}^k(t))$. 
Heat values of different locations are labeled manually based on Explore-Bench data set \cite{xu2022explore}.
Besides, the backbone of this network is selected as the Pointnet \cite{qi2017pointnet}.

Ultimately, since we aim to select a main node that is far away from main nodes in the current FHT-Map while also having a large heat value.
We consider formulating this problem as a multi-objective optimization problem
\begin{equation}
	\label{equ:select_main_opt}
	\begin{aligned}
		\min_{\tau} &\ \{\mathcal{D}(\textbf{p}^k(\tau)),\ -\mathcal{H}(\textbf{p}^k(\tau))\}\\
		s.t.&\ \gamma_2 < \mathcal{D}(\textbf{p}^k(\tau)) <\gamma_1\\
		&\ \boldsymbol{\phi}(\textbf{p}^k(\tau))^T \boldsymbol{\phi}^{k,i}<\theta_{\text{sim}},\ \forall v_f^{k,i} \in \mathcal{V}_{f,\text{main}}^k,
	\end{aligned}
\end{equation}
where $\theta_{\text{sim}}$ is the similarity threshold between two locations \cite{he2022online}.
$\gamma_1$ and $\gamma_2$ are two parameters which control the number of potential main nodes, which is the same as \cite{song2023fht}.

In Eq. \ref{equ:select_main_opt}, the feasible set of $\textbf{p}^k(\tau)$ is determined firstly.
As $\mathcal{R}^k$ explores the environment, it gradually moves towards unknown areas, leading to a reduction in main nodes density $\mathcal{D}(\textbf{p}^k(t))$. 
Besides, when $\mathcal{D}(\textbf{p}^k(t))$ is smaller than $\gamma_1$, the robot will store all features it captures at a fixed rate (30 Hz).
These features are called potential main nodes.
When $\mathcal{D}(\textbf{p}^k(t))$ falls below $\gamma_2$, the single main node selection is performed.
Firstly, potential main nodes are defined as those who satisfy $\gamma_2 < \mathcal{D}(\textbf{p}^k(\tau)) <\gamma_1$.
Then, we do not expect that the selected main node has high similarity with the existing main nodes in the FHT-Map, which will lead to wrong estimations for RP.
Potential main nodes are filtered using $\boldsymbol{\phi}(\textbf{p}^k(\tau))^T \boldsymbol{\phi}^{k,i}<\theta_{\text{sim}}$.
Finally, a new main node will be chosen among the remaining potential main nodes.

Based on Eq. \ref{equ:select_main_opt}, we can obtain a series of Pareto-optimal positions.
However, due to that we want to construct a single main node by solving Eq. \ref{equ:select_main_opt} in a single instance,  weighted sum method is used to turn it into a single objective optimization problem, which is called scalarization
\begin{equation}
	\label{equ:change_form}
	\min_{\tau} \ \omega_1 \mathcal{D}(\textbf{p}^k(\tau)) - \omega_2 \mathcal{H}(\textbf{p}^k(\tau)),
\end{equation}
where $\omega_1,\omega_2 \in \mathbb{R}^+$.
Due to the number of potential main nodes satisfying constraints in Eq. \ref{equ:select_main_opt} is finite, Eq. \ref{equ:select_main_opt} and \ref{equ:change_form} can be solved by enumeration.
Then, the next main node can be established. 
Besides, support nodes can be created automatically based on the criterion in \cite{song2023fht}.
In the following section, $\mathcal{R}^k$ will transmit and receive constructed FHT-Maps among its neighbors $\mathcal{N}_c^k$, thus performing RP estimation.

\subsubsection{RP Estimation}
Assuming that $\mathcal{R}^k$ is located in $\textbf{p}^k(t)$, a visual feature $\boldsymbol{\phi}(\textbf{p}^k(t))$ and a laser scan $\boldsymbol{\psi}(\textbf{p}^k(t))$ can be extracted at this location.
By comparing $\boldsymbol{\phi}(\textbf{p}^k(t))$ and $\boldsymbol{\psi}(\textbf{p}^k(t))$ with information stored in neighbors' FHT-Maps, a single RP estimation can be achieved.
The detailed algorithm is presented in Algorithm \ref{alg:single_est}.

\begin{algorithm}
	\KwIn{robot $\mathcal{R}^k$ at $\textbf{p}^k(t)$, FHT-Maps of $\mathcal{N}_c^k$, visual feature $\boldsymbol{\phi}(\textbf{p}^k(t))$, laser scan $\boldsymbol{\psi}(\textbf{p}^k(t))$} 
	\KwOut{list of estimated RPs $T^k$}
	$T^k=\emptyset$\\
	\For{\rm $\mathcal{G}_f^i \in \mathcal{G}_f(\mathcal{N}_c^k)$}
	{
		\For{\rm $v^{i,j}_{f,\text{main}} \in \mathcal{V}_{f,\text{main}}^i$}
		{
			$r_{\text{match}}$ $\leftarrow$ $\boldsymbol{\phi}(\textbf{p}^k(t))^T \boldsymbol{\phi}^{i,j}$\\
			\If{\rm $r_{\text{match}} > \theta_{\text{match}}$}
			{
				$T^k_i$ $\leftarrow$ \texttt{GICP}($\boldsymbol{\psi}(\textbf{p}^k(t))$, $\boldsymbol{\psi}^{i,j}$)\\
				$T^k \leftarrow T^k \cup T^k_i$\\
			}
		}
	}
	\texttt{broadcast}($T^k$)
	\caption{Single Relative Pose Estimation}
	\label{alg:single_est} 
\end{algorithm}

In Algorithm \ref{alg:single_est}, $\mathcal{R}^k$ compares $\boldsymbol{\phi}(\textbf{p}^k(t))$ with visual features in neighbors' FHT-Map firstly.
If a matched main node is found which means $r_{\text{match}}$ is larger than the threshold $\theta_{\text{match}}$, RP $T^k_i$ between $\mathcal{R}^k$ and $\mathcal{R}^i$ can be obtained using global Iterative Closest Point (ICP).
In our system, there is no central server and each robot makes decisions locally.
Therefore, each robot should have the same global information, for example, estimated RPs ($T^k$), $\mathcal{G}_r$, the pose of each robot in its map frame $\mathbf{P}^k$, and frontiers of each robot.
Since the communication graph $\mathcal{G}_c$ is a connected graph, these lightweight information can be \texttt{broadcast} to all robots.

\subsection{Partition of the Space}
Multiple estimations for RP can be obtained during PIER.
Any edge in $\mathcal{G}_r$ indicates that there is at least one estimated RPs.
Initially, $\mathcal{E}_r$ is an empty set.
When single estimations are gradually added to $\mathcal{E}_r$, $\mathcal{G}_r$ will become partially connected. 
Assuming that $\mathcal{G}_r$ has $m$ connected components, which can be represented by $\mathcal{X}_i, i=1,2\cdots m$.
Based on the definition of connected components, we can be obtained that
\begin{equation}
	\begin{aligned}
		\mathcal{X}_1 \cup \mathcal{X}_2 \cup \cdots \mathcal{X}_m=\mathcal{V}_r\\
		\mathcal{X}_i \cap \mathcal{X}_j = \varnothing ,i \neq j.
	\end{aligned}
\end{equation}
For any two robots in the same connected component, their RP is known. 
If they are not in the same connected component, their RP is unknown.
To obtain the RPs in $\mathcal{X}_i$, the PGO can be performed using
\begin{equation}
	\label{equ:PGO}
	\left\{
	\begin{aligned}
			&\mathop{\min} \limits_{\hat{T}_j,j\in \mathcal{X}_i} \sum_{m,n\in  \mathcal{X}_i} \sum_{k} ||\Omega^{\frac{1}{2}} \epsilon_{mnk}||_2\\
			s.t.\ &\epsilon_{mnk}=[\Delta \theta_{mnk},\Delta x_{mnk},\Delta y_{mnk}]^T,\, 
	\end{aligned}
	\right.
\end{equation}
where $\hat{T_j}$ is the estimated translation matrix of $\mathcal{R}^j$ in a reference frame.
$\Delta \theta_{mnk}$, $\Delta x_{mnk}$, and $\Delta y_{mnk}$ are the rotation and translation components of the matrix $\hat{T}_m^{-1} \hat{T}_n T^n_{m,k}$.
$T^n_{m,k}$ is the k-th estimated RP from $\mathcal{R}^n$ to $\mathcal{R}^m$.
$\Omega \in \mathbb{R}^{3\times 3}$ is information matrix for PGO.
Eq. \ref{equ:PGO} can be solved using Ceres solver.
Then, RPs in $\mathcal{X}_i$ can be obtained.

Collaborative exploration can be performed based on the estimated RPs.
For any connected component $\mathcal{X}_i$, partition of the space is executed among robots in $\mathcal{X}_i$.
Similarly with \cite{bi2023cure}, Voronoi diagram is used for the partition of space and the collaborative behavior of robots is enhanced.
The positions of robots in $\mathcal{X}_i$ are set as base points for this Voronoi diagram.
If $\mathcal{R}^k \in \mathcal{X}_i$, then $S^k$ for $\mathcal{R}^k$ can be obtained by
\begin{equation}
	S^k = \{\mathbf{p}\in \mathbb{R}^2 | D(\mathbf{p},\hat{\mathbf{p}}^k) \leq D(\mathbf{p},\hat{\mathbf{p}}^j),j\neq k,\mathcal{R}^j\in \mathcal{X}_i\},
\end{equation}
where $\mathbf{p}$ is a point in the Voronoi diagram partition for $\mathcal{R}^k$.
$\hat{\mathbf{p}}^k$ is the position of $\mathcal{R}^k$ in the reference frame.
$D(\cdot)$ denotes the distance function between $\mathbf{p}$ and robot position.
In this paper, Euclidean distance $D(\mathbf{p},\hat{\mathbf{p}}^k) = ||\mathbf{p} - \hat{\mathbf{p}}^k||_2$ is used.
The partition of the space also needs to be dynamically adjusted during the exploration. 
We set the criteria for re-partitioning as follows: when new estimations for RP are obtained or the region assigned to any robot has been fully explored.
Besides, considering that each robot has same global information using $\mathcal{G}_c$, for example, estimated RPs, each robot can perform PGO to obtain the refined poses of other robots.
Then, each robot can execute the process of space partition independently and decide where to explore next.

\subsection{Algorithm for PIER}
The overall process for PIER is presented in Algorithm \ref{alg:PIER}.
Before robots start moving, no RPs have been obtained, each robot performs NBV-based exploration.
As robots explore the environment, RP is obtained and the space is partitioned for collaborative exploration.
PIER will be executed continuously until $\mathcal{G}_r$ becomes a connected graph.

\begin{algorithm}
	\KwIn{Graph for RP $\mathcal{G}_r$, communication graph  $\mathcal{G}_c$} 
	\While{\rm \texttt{notConnected}($\mathcal{G}_r$)}
	{
		\For{$\mathcal{R}^k \in \mathcal{R}$}
		{
			\If{\rm \texttt{UpdatePartition}()}
			{
				$\mathcal{X}(\mathcal{R}^k)$ $\leftarrow$ \texttt{FindComponent}($\mathcal{R}^k$)\\
				$S^k$ $\leftarrow$ \texttt{VoronoiPartition}($\mathcal{R}^k$,$\mathcal{X}(\mathcal{R}^k)$)\\
				$\{\hat{f}^k\}$ $\leftarrow$  \texttt{ObtainFrontiers}($S^k$)\\
				\If{\rm \textbf{Len}($\{\hat{f}^k\}$)=0}
				{
					\texttt{ReAssign}($\mathcal{R}^k$)\\
				}
			}
			\Else
			{
				\texttt{NBVExploration}()\\
				\texttt{RPEstimation}($\mathcal{N}_c^k$)
			}
		}
	}
	\caption{PIER Implementation}
	\label{alg:PIER} 
\end{algorithm}

Some key functions in Algorithm \ref{alg:PIER} are explained as follows.
\begin{itemize}[]
	\item \texttt{UpdatePartition()}: When new estimations for RP are obtained or the region assigned to a robot has been fully explored, this function will return True.
	\item \texttt{FindComponent}($\mathcal{R}^k$): For each robot $\mathcal{R}^k$, find the connected component $\mathcal{X}(\mathcal{R}^k)$ that contains $\mathcal{R}^k$.
	\item \texttt{VoronoiPartition}($\mathcal{R}^k$,$\mathcal{X}(\mathcal{R}^k)$): For each robot $\mathcal{R}^k$, perform partition of the space based on Voronoi diagram.
	\item \texttt{ReAssign}(): For $\mathcal{R}^k$ with no frontiers in $S^k$, select the frontier furthest from any robot in  $\mathcal{X}(\mathcal{R}^k)$ as its target.
	\item \texttt{RPEstimation}($\mathcal{N}_c$): For each robot, Algorithm \ref{alg:single_est} is performed continuously to obtain RP.
\end{itemize}

\section{Optimal Rendezvous Point Selection}
\label{sec:OptRend}
When single estimations are gradually added to $\mathcal{E}_r$, $\mathcal{G}_r$ will become connected at time $t_0$. 
We assume that the pose for $\mathcal{R}^k$ in a reference frame at $t_0$ is $\tilde{\textbf{p}}^k(t_0)$.
At this point, each robot broadcasts its own FHT-Map with only structural information.
The environmental features $\boldsymbol{\phi}^{k,i}$ and $\boldsymbol{\psi}^{k,i}$ with large storage load are not transmitted.
Then, FHT-Maps are merged using estimated RPs, where two situations are considered.
Firstly, an edge is added between two main nodes if there exists an estimated RP among them.
Secondly, if $\textbf{p}^{k',j}\in s^{k,i}$, which indicates that node $v_f^{k',j}$ lies in the local free space of node $v_f^{k,i}$, an edge between these two nodes will be created.
The goal for selecting an optimal rendezvous point is to find a point $\textbf{p}^*$ where all robots can converge as quickly as possible.
This problem can be formulated as minimizing the maximum path length from each robot to point $\textbf{p}^*$
\begin{equation}
    \label{min_max}
		\begin{aligned}
		&\min_{\textbf{p}^*} \max_{k=1\cdots n} D(\tilde{\textbf{p}}^k(t_0),\textbf{p}^*)\\
		& s.t.\ \textbf{p}^* \in S_{\text{free}},
		\end{aligned}
\end{equation}
where $D(\tilde{\textbf{p}}^k(t_0),\textbf{p}^*)$ is the length of the shortest path from $\tilde{\textbf{p}}^k(t_0)$ to $\textbf{p}^*$.
Since $S_{\text{free}}$ is a non-convex space, the problem is not convex. 
Besides, due to only a sparse representation of the environment is used in this work, we can only present an algorithm that finds a sub-optimal solution $\hat{\textbf{p}}^*$ compared to the global optimal value.

In the FHT-Map, each node $v_f^{k,i}$ stores a convex local free space $s^{k,i}$.
$\hat{\textbf{p}}^*$ may exist in any of the $s^{k,i}$.
Besides, since the problem is non-convex, we solve it by an approach based on sampling.
Points can be sampled in each $s^{k,i}$ with a density of $\delta$.
Then, based on path planning methods on the FHT-Map \cite{song2023fht}, we can obtain the value $F(\cdot)=\max_{k=1\cdots n} D(\tilde{\textbf{p}}^k(t_0),\cdot)$ for each point.
Finally, the minimal $F(\cdot)$ of sampled points will be selected as $\hat{\textbf{p}}^*$.
The algorithm for selecting sub-optimal rendezvous point $\hat{\textbf{p}}^*$ is presented in Algorithm \ref{alg:opt_p} based on the merged topological map.

\begin{algorithm}
	\KwIn{poses of all nodes in FHT-Maps $\{\textbf{p}^{k,i}\}$, local free spaces of all nodes $\{s^{k,i}\}$, edges in all FHT-Maps $\{e_f^{k,i}\}$, the density for sampling $\delta$}
	\KwOut{sub-optimal rendezvous point $\hat{\textbf{p}}^*$}
	$\{F(\textbf{p}^{k,i})\}$ $\leftarrow$ \texttt{CalculateF}($\{\textbf{p}^{k,i}\}$,$\{s^{k,i}\}$,$\{e^{k,i}\}$)\\
	$L= \{F(\textbf{p}^{k,i})\}$\\
	\For{\rm $s^{k,i} \in \{s^{k,i}\}$}
	{
		PointSet $\leftarrow$ \texttt{Sample}($s^{k,i}$,$\delta$)\\
		$L'$ $\leftarrow$ \texttt{CalculateF}($\mathbf{p}^\text{PointSet}$,$\{s^{k,i}\}$,$\{e^{k,i}\}$)\\
		$L \leftarrow L \cup L'$\\
	}
	OptIndex $\leftarrow \mathop{\arg\min} L$, $\hat{\textbf{p}}^* \leftarrow \mathbf{p}^{\text{OptIndex}}$\\
	\Return{$\hat{\textbf{p}}^*$}
	\caption{Optimal Rendezvous Point Selection}
	\label{alg:opt_p} 
\end{algorithm}
Due to that \texttt{CalculateF}() (\textbf{Line} 1 and 5), which means calculate $F(\cdot)$ for given points, involves virtual path planning on the topological maps, it is the main time-consuming step in Algorithm \ref{alg:opt_p}.
However, the time spent on \texttt{CalculateF}() can be reduced.
Firstly, path length from $\tilde{\textbf{p}}^k(t_0)$ to each node $\{\textbf{p}^{k,i}\}$ (\textbf{Line} 1) can be quickly obtained in the FHT-Map.
Besides, path planning for points $\mathbf{p}^\text{PointSet}$ in $s^{k,i}$ (\textbf{Line} 5) consists of path on FHT-Map and path from nodes to $\mathbf{p}^\text{PointSet}$, where any actual path planning algorithm is not required.
So, our proposed method is very time-efficient.

\section{Simulations and Experiments}
\subsection{Experimental Setup}
\subsubsection{Simulation Setup}
Simulations are conducted in the ROS-Gazebo simulator with AMD Ryzen 3900X CPU at 4.00\,GHz (24 cores), 64\,GB RAM, and NVIDIA RTX 2080 Ti GPU. 
Turtlebot Burger robots are used in this work, and each robot is equipped with an RPLIDAR A2 LIDAR (sensing range 12\,m), which is the same for real-world experiments.
Cartographer is used for localization during exploration.
The robot uses four RGB cameras which are arranged horizontally in a circular manner without overlap of Field of View (FOV) to capture information from all directions.
Each camera has a FOV of $90^{\circ}$, with a resolution of 640$\times$480.
Two simulation environments are used in this work, where \textit{small} is 485\,$\text{m}^2$ and \textit{office} is 3285.22\,$\text{m}^2$.
In all simulations, $\gamma_1$ and $\gamma_2$ are set to $\exp(-0.5)$ and $\exp(-1)$, $\omega_1$ and $\omega_2$ are set to 1 and 2, $\theta_\text{sim}$ is set to 0.94, $\theta_{\text{match}}$ is 0.96, and $\sigma_c$ is set to 4.
\subsubsection{Real-world Experiment}
To validate the effectiveness of our method, real-world experiments are conducted in an indoor environment with size 5.5 $\times$ 9.3\, m$^2$ using two robots.
Turtlebot Burger robots equipped with a NVIDIA Jetson TX2 and a Realsense D435i camera are used in the experiments.
Besides, due to the size of the real-world scene is small, to guarantee that enough number of main nodes can be established for RP estimation,  we set $\sigma_c$ to 1, while keeping other hyper-parameters constant.

\subsection{Case Study}
\begin{figure*}[!t]
	\centering
	\includegraphics[width=7in]{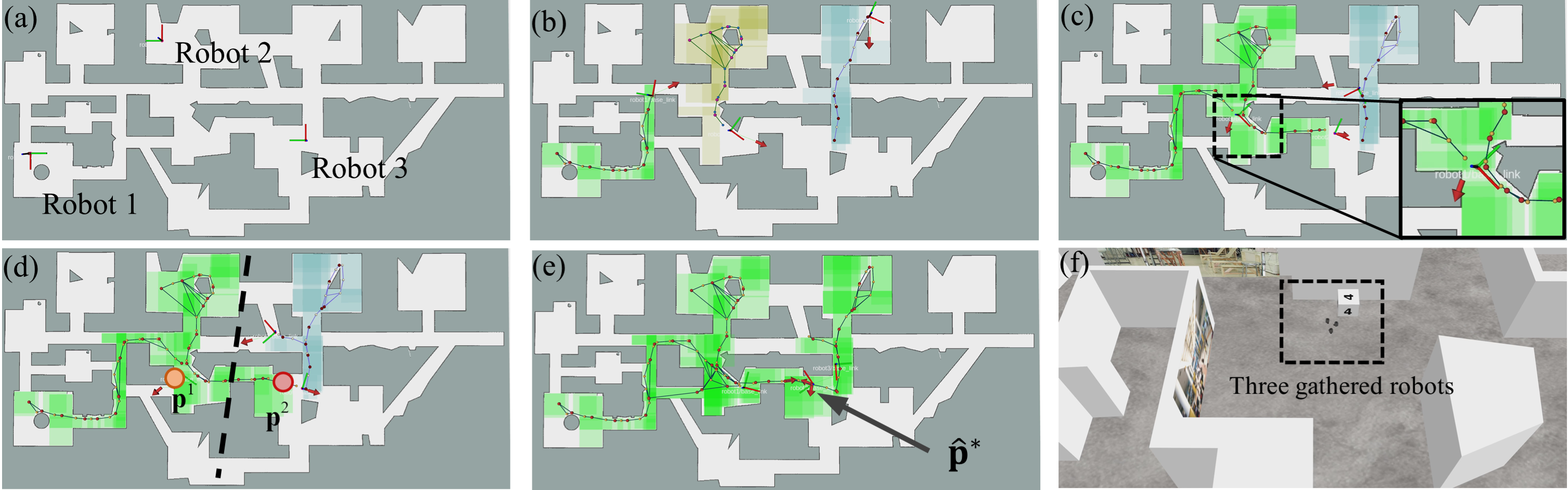}
	\caption{
		The rendezvous of three robots in office Environment.
		(a) Three robots are located in an unknown environment.
		(b) Each robot perform PIER to explore the environment, while constructing a FHT-Map.
		(c) $\mathcal{R}^1$ passes through a main node of $\mathcal{R}^2$, thereby merging two topological maps $\mathcal{G}_{f}^1$ and $\mathcal{G}_{f}^2$.
		(d) Space is partitioned between $\mathcal{R}^1$ and $\mathcal{R}^2$. Then $\mathcal{R}^2$ finds a main node of the remaining  $\mathcal{R}^3$.
		(e) Rendezvous point $\hat{\textbf{p}}^*$ is selected using the merged FHT-Map. Then each robots performs rendezvous.
		(f) Rendezvous among three robots is realized in the simulation.
	} 
	\label{fig:rend_vis}
\end{figure*}
In the first simulation, three robots are located in the \textit{office} environment with initial poses (5\,m, 15\,m, -90$^\circ$), (26\,m, 33\,m, 90$^\circ$), and (49\,m, 17\,m, 90$^\circ$).
The communication graph $\mathcal{G}_c$ is defined as a fully connected graph.
Besides, each robot is unaware of the positions of others.
The detailed procedure for rendezvous is shown in Fig. \ref{fig:rend_vis}.

Firstly, each robot performs exploration of the environment, concurrently constructing FHT-Maps of their observed areas. 
At $t=$106\,s, $\mathcal{R}^1$ moves near the main node of $\mathcal{R}^2$ and obtains the RP between the two robots. 
At this point, both two robots execute the PIER strategy. 
At $t=$126\,s, $\mathcal{R}^2$ observes a main node constructed by $\mathcal{R}^3$, thus making $\mathcal{G}_r$ a connected graph.
At $t=$201\,s, rendezvous is achieved among these three robots.
Throughout this process, the total amount of transmitted data is 164.69\,KB, indicating that multi-robot rendezvous is achieved with minimal communication burden.

\subsection{Comparison and Ablation Studies}
\label{sec:com_eval}
To validate the proposed method, comparison and ablation studies are conducted.
We compare our overall framework with two methods.
The first one transfers \textbf{g}rid \textbf{m}aps (\textbf{GM}) between robots and the second one transfers traditional \textbf{t}opological \textbf{m}aps (\textbf{TM}).
In \textbf{GM}, the next goal for $\mathcal{R}^k$ is selected as the frontier among all frontiers in $\mathcal{X}(\mathcal{R}^k)$ with maximum utility based on \cite{umari2017autonomous}.
RP estimation is achieved based on ROS package Multirobot\_Map\_Merge.
Besides, 10 points are randomly sampled in the merged map and the point with minimal $F(\cdot)$ is selected for rendezvous.
In \textbf{TM}, the robot exploration and RP estimation is based on traditional topological maps \cite{zhang2022mr}, and the rendezvous point is selected among all nodes.
In these methods, once the RPs between all robots are determined, rendezvous will be performed.
The time when exploration is finished is defined as $\mathbf{t_0}$, the time when rendezvous is achieved is defined as $\mathbf{t_1}$, and the volume of transmitted data is also presented in Table \ref{table:compare_exp}.

Compared with \textbf{GM}, we reduce $\mathbf{t_0}$ by 29.15\,\% and 62.11\,\% in \textit{small} and \textit{office} respectively.
Besides, due to only a lightweight topological map is transmitted in this work, we reduce the communication burden by 91.50\,\% and 96.02\,\%.
Compared with \textbf{TM}, we reduce the time for achieving rendezvous $\mathbf{t_1}$ by 46.43\,\% and 33.99\,\%. 
It can be attributed to that PIER and transferred $\boldsymbol{\psi}^{k,i}$ can accelerate the acquisition of RP.
And the more detailed representation of the environmental structure allows for a better selection rendezvous point.
Besides, the average data upload per robot per second is 0.493\,KB based on FHT-Map in \textit{office}. 
With the bandwidth of 1\,Mbps, the latency introduced is 3.943 milliseconds, which has minor influence on the system.

Ablation studies are conducted to validate the effectiveness of the exploration strategy and the proposed main node selection module.
As for exploration strategy, we compare PIER against RRT-based \cite{umari2017autonomous} (\textbf{RRT}), which is the same as \textbf{GM}, and m-TSP-based \cite{hardouin2023multirobot} (\textbf{TSP}) exploration.
Detailed results are presented in Table \ref{table:compare_exp}.
PIER outperforms the other two methods in terms of exploration time and transmitted data. 
Compared with \textbf{RRT} and \textbf{TSP}, PIER reduces the $\mathbf{t_0}$ by 30.32\,\% and 49.08\,\% in \textit{small} (29.44\,\% and 41.20\,\% in \textit{office}).
Due to exploration time is reduced, a smaller map is constructed, which also results in a decrease in the amount of transmitted data.
Besides, we remove heat value evaluation in main nodes selection (\textbf{NS}), and only $\mathcal{D}(\textbf{p}^k(\tau))$ is considered in Eq. \ref{equ:select_main_opt}.
The introduction of heat value evaluation can accelerate the exploration by 13.47\% and 16.83\% in two environments.

\begin{table}[!t]
	\fontsize{8}{7}\selectfont
	\renewcommand{\arraystretch}{1.4} 
	\setlength\tabcolsep{2pt}   
	\caption{Comparison and ablation studies. Methods with ``*'' and ``\#'' are comparison and ablation experiments respectively.}
	\label{table:compare_exp}
	\centering
	\begin{tabular}{C{0.8cm}C{0.8cm}C{1cm}C{0.9cm}C{1cm}C{0.9cm}C{1cm}C{0.9cm}}
		\toprule
		\toprule
		\multicolumn{1}{c}{\footnotesize \textbf{Scene}} &\multicolumn{1}{c}{\footnotesize \textbf{Method}} & \multicolumn{2}{c}{\footnotesize $\mathbf{t_0}$ (s)}  &  \multicolumn{2}{c}{\footnotesize $\mathbf{t_1}$ (s)} & \multicolumn{2}{c}{\footnotesize Volume (KB)}\\
		\cmidrule(lr){3-4}\cmidrule(lr){5-6}\cmidrule(lr){7-8}
		&  & \textbf{Avg} & \textbf{Std} &\textbf{Avg} &\textbf{Std}& \textbf{Avg} & \textbf{Std}\\
		              & \textbf{GM}* &  74.62	&24.62 &124.949& 45.05 &976.41	&49.90\\
		              & \textbf{TM}* &  95.71	&22.78 &149.19&	37.53 &134.52	&26.13\\
		\textbf{Small}& \textbf{RRT}\# &  75.44	&25.48 &111.18&	40.33 &97.30	&37.31\\
			          & \textbf{TSP}\# &  103.24&17.76 &135.43&	20.73 &104.18	&16.40\\
		              & \textbf{NS}\# & 60.75 &\textbf{9.32} &101.02&	32.07         &88.00	&15.79\\
		              & \textbf{Ours} & \textbf{52.57}&	11.33&	\textbf{79.92}&	\textbf{13.61} &\textbf{82.95}&	\textbf{5.88}\\
		\cmidrule(lr){1-8}
		            & \textbf{GM}* &  392.94	&141.93 &537.53& 160.61 &5531.84 &782.55\\
					& \textbf{TM}* &  195.70	&\textbf{20.48} &314.31&	\textbf{37.53} &365.72	&\textbf{60.06}\\
	 \textbf{Office}& \textbf{RRT}\# &  211.03	&69.98 &303.05&	71.61 &305.46	&121.06\\
					& \textbf{TSP}\# &  253.22&41.23 &357.62&	45.09 &247.82	&81.51\\
					& \textbf{NS}\# & 179.04 &46.46 &253.66&	101.39 &\textbf{206.45}	&62.92\\
					& \textbf{Ours} & \textbf{148.90}&	41.93&	\textbf{207.48}&49.38 &220.14&	66.59\\
		\bottomrule
		\bottomrule
	\end{tabular}
\end{table}

Three hyper-parameters $\gamma_1$, $\gamma_2$, and $\sigma_c$ are used in the main node selection of experiments.
A smaller $\sigma_c$ or a larger $\gamma_2$ will contribute to an FHT-Map with more main nodes.
It can improve the probability of finding matched main nodes, but it leads to an increase in communication burden.
$\gamma_1$ influences the ratio of potential main nodes.

\subsection{Evaluation on Rendezvous Point Selection}
\begin{figure*}[!t]
	\centering
	\includegraphics[width=7in]{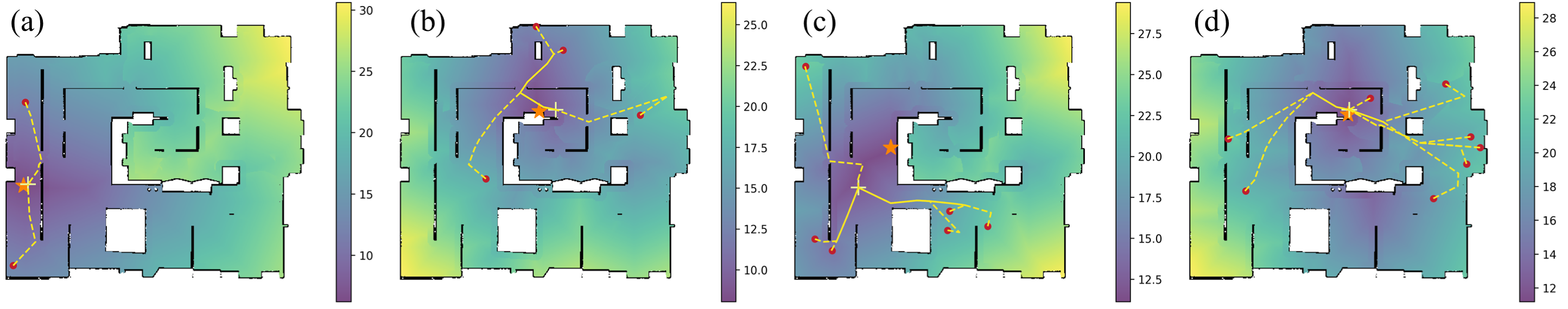}
	\caption{Illustration of the selected rendezvous points. The color in the map represents $F(\textbf{p})$ at location $\textbf{p}$ obtained by A* algorithm.
		The red circles represent robots' positions. 
		The yellow star indicates the global optimal points $\textbf{p}^*$. 
		The plus sign represents the position of $\hat{\textbf{p}}^*$ based on FHT-Maps.
		(a) Two robots. (b) Four robots. (c) Six robots. (d) Eight robots.}
	\label{fig:Rendpoint_vis}
\end{figure*}

Due to that the rendezvous point $\hat{\textbf{p}}^*$ we selected is based on the topological map, so $\hat{\textbf{p}}^*$ is not the global minimal point $\textbf{p}^*$. 
In this section, we compared $\hat{\textbf{p}}^*$ with the globally optimal $\textbf{p}^*$ based on the grid map.
The results of different numbers of robots are illustrated in Fig. \ref{fig:Rendpoint_vis}.
The distance errors $||\hat{\textbf{p}}^* - \textbf{p}^*||_2$ between $\hat{\textbf{p}}^*$ and $\textbf{p}^*$ are 0.364, 1.25, 3.83, and 0.364\,m, respectively.
As for the error in path length $F(\hat{\textbf{p}}^*)-F(\textbf{p}^*)$, they are 0.243, 1.99, 0.925, and 0.738\,m.
It indicates that the rendezvous point estimated based on FHT-Maps is close to the global optimal solution.

\begin{figure}[!t]
	\centering
	\includegraphics[width=3.5in]{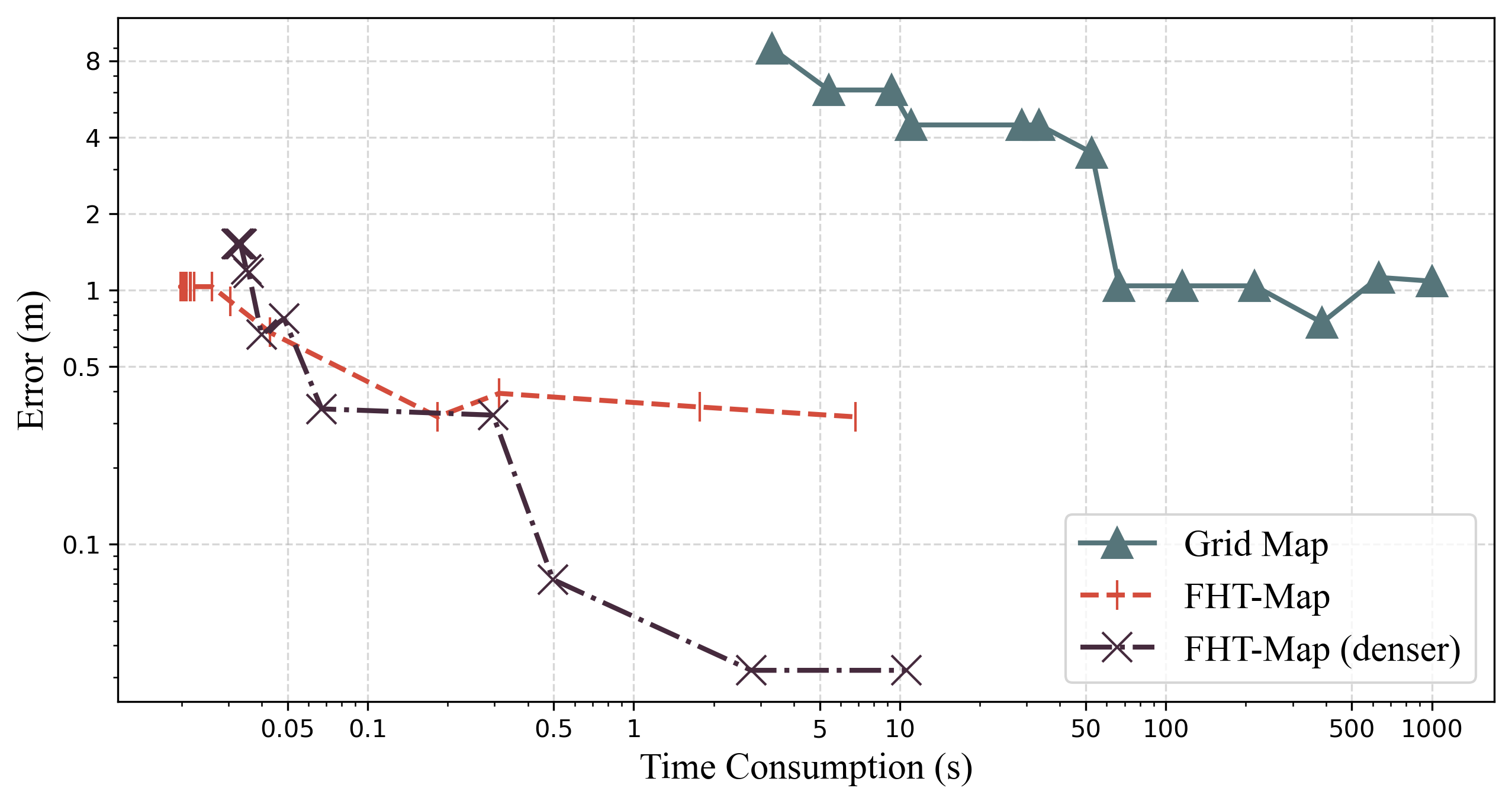}
	\caption{
		Comparison on time consumption and error based on different map.
		A near optimal solution $\hat{\textbf{p}}^*$ can be obtained on FHT-Map.
		Besides, if there are denser nodes in FHT-Map, a more accurate result can be obtained.} 
	\label{fig:opt_p_time_compare}
\end{figure}

We also compare the efficiency of rendezvous point computation based on the topological map and the grid map.
For our method, the computation time is influenced by the sampling density $\delta$. 
For the method based on the grid map, we randomly sample $m$ points in free space and use the A* algorithm to obtain $F(\cdot)$, and the point with minimum $F(\cdot)$ is selected as the rendezvous point. 
The main factor affecting the computation time is $m$.
Fig. \ref{fig:opt_p_time_compare} reports the relationship between computation time and error $F(\hat{\textbf{p}}^*)-F(\textbf{p}^*)$.
For the FHT-Map with support node density of 2\,m (red line), compared to the grid map, it achieves a result closer to the optimal solution at a low computation time. 
However, as $\delta$ decreases further, it cannot obtain a more accurate result. 
It is because the topological map is a sparse representation of the environment, which prevents obtaining accurate shortest distances between two points.
However, by using an FHT-Map with denser nodes (node density is 1\,m, brown line), we are able to obtain a result that is closer to the theoretically optimal rendezvous point.

\subsection{Real-world Experiment}
\begin{figure}[!t]
	\centering
	\includegraphics[width=3.3in]{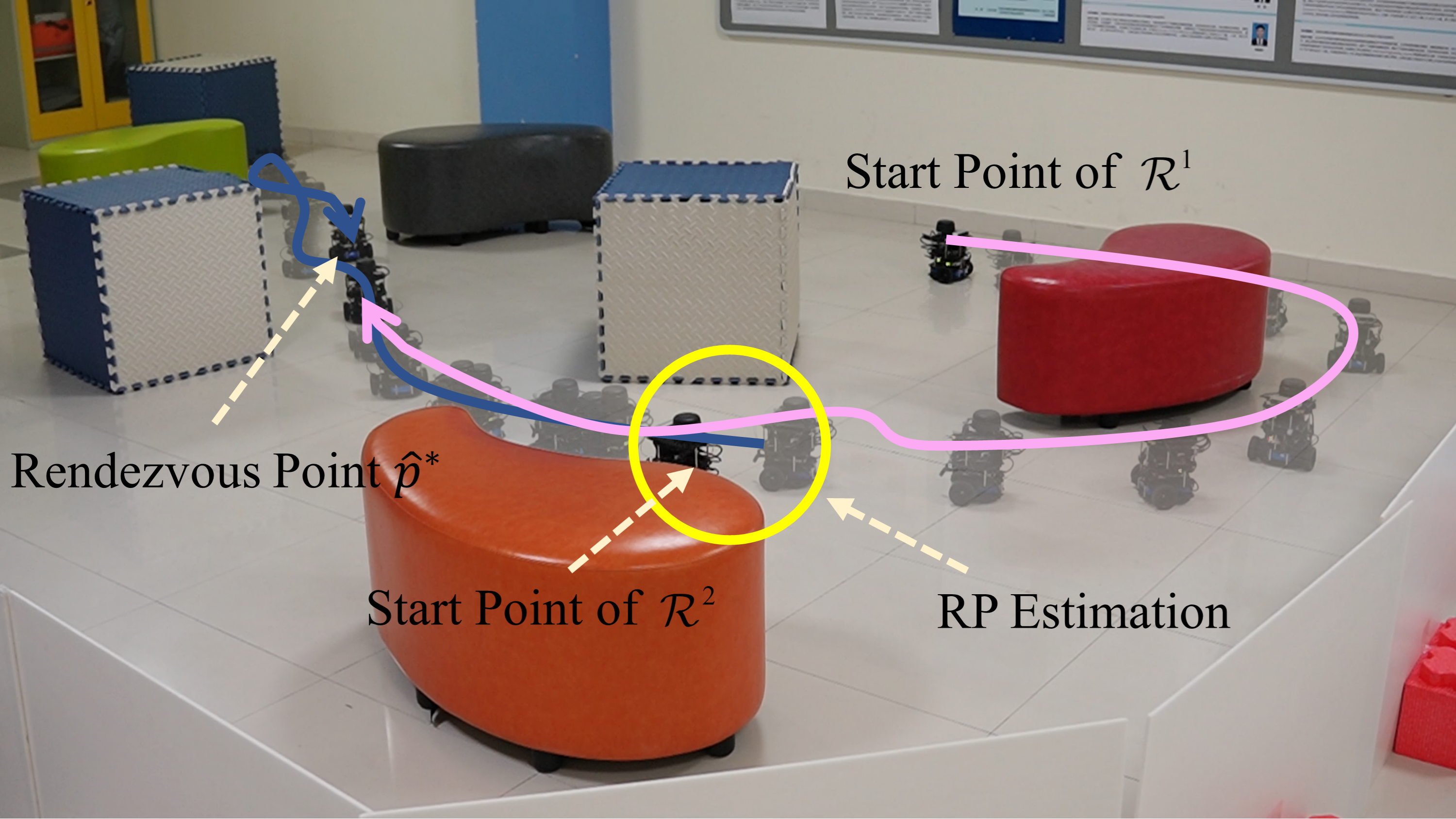}
	\caption{
		The real-world experiment of two robots.
		The pink and blue arrows represent trajectories of of two robots.
	}
	\label{fig:real_world}
\end{figure}

A snapshot of the real-world experiment is shown in Fig. \ref{fig:real_world}.
It takes 91\,s for PIER and 57\,s to achieve rendezvous.
WiFi is used for transmitting data.
The total amount of transmitted data is 26.71\,KB, which indicates that rendezvous is achieved within constrained of communication bandwidth.

\section{Conclusion \& Future Work}
In this work, we present a solution for the rendezvous problem in unknown environments with limited communication.
Firstly, PIER is proposed for the task of incomplete exploration.
Then, each robot constructs and shares an FHT-Map during exploration, and RP estimation can be realized based on this map.
Finally, when $\mathcal{G}_r$ becomes a connected graph, a sub-optimal rendezvous point will be selected.
Simulations are conducted to validate the effectiveness of the proposed method, where a shorter time for exploration and calculation of $\hat{\textbf{p}}^*$ is achieved.

The main limitation of the proposed method lies in the lack of consideration for potential erroneous results in RP estimations, which may lead to the failure of rendezvous. 
In future work, we aim to incorporate this aspect into our work and conduct more physical experiments, as well as implement rendezvous strategies in more challenging environments.

\bibliographystyle{ieeetr} 
\bibliography{ref}
\end{document}